\documentclass[conference]{IEEEtran}
\IEEEoverridecommandlockouts

\usepackage{amsthm}
\usepackage{cite}
\usepackage{amsmath,amssymb,amsfonts}
\usepackage{algorithmic}
\usepackage{graphicx}
\usepackage{textcomp}
\usepackage{xcolor}
\usepackage{import}

\usepackage{amsthm}

\DeclareMathOperator{\w}{\boldsymbol{w}}
\DeclareMathOperator{\x}{\boldsymbol{x}}

\begin{document}
\setlength{\textfloatsep}{6pt}

\title{Convergence Analysis of alpha-SVRG \\ under Strong Convexity
\thanks{The work of S. Park and S. Vlaski was supported by EPSRC Grants EP/X04047X/1 and EP/Y037243/1. Emails: \{ sean.xiao20, s.park, s.vlaski \}@imperial.ac.uk}
}

\author{\IEEEauthorblockN{Sean Xiao, Sangwoo Park and Stefan Vlaski}
\IEEEauthorblockA{\textit{Department of Electrical Engineering, Imperial College London}} 
}

\newtheorem{corollary}{Corollary}
\newtheorem{definition}{Definition}
\newtheorem{proposition}{Proposition}
\newtheorem{lemma}{Lemma}
\newtheorem{theorem}{Theorem}
\newtheorem{assumption}{Assumption}

\maketitle

\begin{abstract}
Stochastic first-order methods for empirical risk minimization employ gradient approximations based on sampled data in lieu of exact gradients. Such constructions introduce noise into the learning dynamics, which can be corrected through variance-reduction techniques. There is increasing evidence in the literature that in many modern learning applications noise can have a beneficial effect on optimization and generalization. To this end, the recently proposed variance-reduction technique, $\alpha$-SVRG \cite{yin2023coefficient} allows for fine-grained control of the level of residual noise in the learning dynamics, and has been reported to empirically outperform both SGD and SVRG in modern deep learning scenarios. By focusing on strongly convex environments, we first provide a unified convergence rate expression for $\alpha$-SVRG under fixed learning rate, which reduces to that of either SGD or SVRG by setting $\alpha=0$ or $\alpha=1$, respectively. We show that $\alpha$-SVRG has faster convergence rate compared to SGD and SVRG under suitable choice of $\alpha$. Simulation results on linear regression validate our theory.
\end{abstract}


\section{Introduction}
We consider the following empirical risk minimization problem:
\begin{align} \label{eq:objective}
    w^o = \arg\min_{w} J(w)= \frac{1}{N} \sum_{n=1}^N Q(w; x_n),
\end{align}
where the finite-sum objective $J(w)$ is a strongly convex function that consists of real-valued loss functions $Q(w;x_n)$ evaluated using the realization $x_n$ given the $M$-dimensional model $w \in \mathbb{R}^M$.

The optimal model \( w^o \) in~\eqref{eq:objective} can be pursued via the stochastic gradient algorithm:
\begin{align}\label{eq:sgd}
    \w_i = \w_{i-1} - \mu_{i-1} \widehat{\nabla J}(\w_{i-1})
\end{align}
where \( \widehat{\nabla J}(\w_{i-1}) \) denotes a stochastic approximation of the gradient \( \nabla J(\w_{i-1}) \) and \( \mu_{i-1} > 0\) is the learning rate. A classical choice is the construction $\widehat{\nabla J}(w) = Q(w;\boldsymbol{x}_i)$ obtained by randomly sampling a single data example $\boldsymbol{x}_i$ from the set \( \{ x_n \}_{n=1}^N \) or its mini-batch variants~\cite{sayed2022inference,simeone2022machine}. We recover gradient descent by setting \( \widehat{\nabla J}(\w_{i-1}) = \nabla J(\w_{i-1}) \). We utilize bold font in~\eqref{eq:sgd} to emphasize the fact that the iterates \( \w_{i} \) are random as a result of the variance introduced by the stochastic approximation \( \widehat{\nabla J}(\w_{i-1}) \). As a result of this persistent noise, stochastic gradient algorithms converge to a small neighborhood of \( w^o \) when constant learning rates are employed~\cite{Yuan16, sayed2022inference}.

To this end, variance reduction methods employing control variates \cite{ross2022simulation} have been proposed \cite{johnson2013accelerating, Defazio14, zhang2013linear, mahdavi2013mixed, wang2013variance, nguyen2017sarah, fang2018spider, wang2019spiderboost} that \emph{mix} the high-variance stochastic gradient with another random vector for which the expectation is known in advance. In this work, we particularly focus on the popular \emph{stochastic variance reduced gradient} (SVRG) \cite{johnson2013accelerating} that utilizes full gradient information obtained in a periodic manner.

Specifically, denoting as $i=1,2,...$ the iteration index, SVRG periodically evaluates the full gradient given every $m$ iterations. Denoting as $\bar{\w}_i$ the \emph{snapshot} model that has been used for such full gradient evaluation, i.e.,  $\bar{\w}_i = \w_{km}$ for $i=km, ..., km+m-1$, SVRG updates the model $\w_i$ by
\begin{equation} \label{eq:svrg_update}
    \w_{i} = \w_{i-1} - \mu_{i-1}\Big(\widehat{\nabla J}({\w}_{i-1}) -  \widehat{\nabla J}(\bar{\w}_{i-1}) + \nabla J(\bar{\w}_{i-1})\Big),
\end{equation}
denoting as $\mu_{i-1}$ the learning rate.

By reducing the variance of the stochastic gradient, it has been shown that SVRG can enhance the convergence rate in both convex \cite{johnson2013accelerating} and non-convex scenarios \cite{reddi2016stochastic}. We refer the reader to \cite{reddi2016stochastic} for details as well as for other variance reduction techniques. Follow-up works have focused on alleviating the requirement of full gradient computation (see, e.g., \cite{frostig2015competing, lei2017less, cutkosky2019momentum})

Despite the clearly documented benefit of variance reduction for optimization, it has been observed that in modern learning applications it may result in reduced performance~\cite{Defazio19}. Motivated by these observations, $\alpha$-SVRG was proposed in \cite{yin2023coefficient}, that adjusts the \emph{variance reduction} level of SVRG by introducing the additional parameter $\alpha\in[0,1]$ that dictates the reliance on the snapshot model. 

In particular, the authors in \cite{yin2023coefficient} propose the following update rule
\begin{align} \label{eq:alpha_SVRG}
    &\w_{i} = \w_{i-1} - \\&\quad \mu_{i-1}\Big(\widehat{\nabla J}({\w}_{i-1}) - \alpha \widehat{\nabla J}(\bar{\w}_{i-1}) + \alpha \nabla J(\bar{\w}_{i-1})\Big). \nonumber 
\end{align}
While the empirical benefit of this construction in deep learning applications was clearly demonstrated in~\cite{yin2023coefficient}, an analytical understanding of this phenomenon is outstanding.

In this work we provide an analytical quantification of the convergence rate of \( \alpha \)-SVRG in strongly-convex environments. The unified expression recovers those of SGD and SVRG by setting $\alpha=0$ and $\alpha=1$, respectively.  Furthermore, our result is general enough to interpolate between these two edge cases thus providing a pathway to choose $\alpha$ as a function of the randomness in $\widehat{\nabla J}(w)$.



\section{Preliminaries}
In this section, we briefly summarize classical convergence results for SGD and SVRG adapted from \cite{sayed2022inference}, which will serve as benchmarks for the performance of \( \alpha \)-SVRG further ahead.

\subsection{Regularity Conditions}
Our analysis will rely on common smoothness and convexity conditions.
\begin{assumption}[Lipschitz gradient of the loss]\label{cond:1} \label{assump:Lip} The gradients of the loss $Q(w; x_n)$ are Lipschitz for every sample $x_n$, i.e., for each $x_n$ and for any pair of models $w_1$ and $w_2$, we have
\begin{align}
    ||\nabla Q(w_1;x_n) - \nabla Q(w_2;x_n)|| \leq \delta_n||w_1-w_2||. 
\end{align}
We will further denote as $\delta^2$ the mean of the squared Lipschitz constants $\delta_n$, i.e.,  $\delta^2 = (1/N)\sum_{n=1}^N \delta_n^2$.
\end{assumption}

\begin{assumption}[$\nu$-strong convexity of the risk]\label{cond:2} \label{assump:convexity}The risk $J(w)$ is $\nu$-strong convex, i.e., it satisfies 
\begin{align}
    J(w) \geq J(w^o) + \nabla J(w^o)^\top (w-w^o) + \frac{\nu}{2}||w-w^o||^2.
\end{align}
\end{assumption}

\subsection{Gradient Noise Bounds}
Under Conditions~\ref{cond:1} and~\ref{cond:2} it follows that the variance induced by the stochastic gradient approximation can be bounded as follows.
\begin{lemma}[Variance bound of SGD gradient noise~\cite{sayed2022inference}] \label{lemma:MSE_SGD}
The expected squared deviation between the risk gradient $\nabla J(w)$ and the stochastic gradient $\widehat{\nabla J}(w) = \nabla Q(w;\x_i)$ with $\x_i$ chosen among $\{x_1,...,x_N\}$ uniformly at random, has the following upper bound for any fixed $w$:
\begin{align}
    \mathbb{E}||\nabla J(w) - \widehat{\nabla J}(w)||^2  \leq 6\delta^2 ||w^o - w||^2 + 3 \sigma^2
\end{align}    
where $\sigma^2$ is the variance of the gradient evaluated at the optimal model $w^o$, i.e.,  $\sigma^2 = (1/N)\sum_{n=1}^N ||\nabla Q(w^o;x_n)||^2$.
\end{lemma}

\begin{lemma}[Variance bound of SVRG gradient noise~\cite{sayed2022inference}] \label{lemma:MSE_SVRG}
The expected squared deviation between the risk gradient $\nabla J(w)$ and the SVRG gradient $\widehat{\nabla J}^\text{SVRG}(w) = \nabla J(\bar{w}) - \nabla Q(\bar{w};\x_i) + \nabla Q(w;\x_i)$ given the snapshot model $\bar{w}$ admits the following bound for any fixed $w$ and \( \overline{w} \):
\begin{align}
    &\mathbb{E}||\nabla J(w) - \widehat{\nabla J}^\text{SVRG}(w)||^2 \nonumber\\&\leq 8\delta^2 \big(||w^o - w||^2 + ||w^o - \bar{w}||^2\big)
\end{align}    
\end{lemma}

\subsection{Convergence Analysis}
Together with the regularity conditions, the gradient noise bounds give rise to the following known convergence guarantees for SGD and SVRG.

\begin{theorem}[Convergence of SGD~\cite{sayed2022inference}] \label{theorem:SGD}
Under the choice of the learning rate $\mu < \nu/7\delta^2$, SGD admits
    \begin{align}
        \mathbb{E}[||w^o-\w_{i}||^2] \leq (1-\mu\nu)^i ||w^o-w_0||^2 + \frac{3\mu\sigma^2}{\nu}.
    \end{align}
\end{theorem}

\begin{theorem}[Convergence of SVRG~\cite{sayed2022inference}] \label{theorem:SVRG}
Under the choice of the learning rate $\mu < \min\{ \nu/9\delta^2, 1/m\nu \}$; of the snapshot period $m>16\delta^2/\nu^2$,  SVRG admits the following for $i=k\cdot m$
\begin{align} \label{eq:SVRG_conv}
    \mathbb{E}[||w^o-\w_{i}||^2] &\leq \Big( (1-\mu\nu)^m  + \frac{8\mu\delta^2}{\nu}\Big)^{k} ||w^o-w_0||^2 
    \nonumber\\&\leq \Big(1-\frac{1}{2}\mu\nu m + \frac{8\mu\delta^2}{\nu}\Big)^{k} ||w^o-w_0||^2.
\end{align}
\end{theorem}

\subsection{Iteration Complexity}
In order to account for differing rates of convergence and steady-state behavior, we will be comparing the performance of SGD, SVRG as well as \( \alpha \)-SVRG further ahead in terms of the iteration complexity, defined as:
\begin{align} \label{eq:def_iteration_complexity}
i^o = \min\{i: \mathbb{E}||w^o-\w_i||^2\leq \epsilon\}.
\end{align}

\begin{corollary}[Iteration complexity of SGD] \label{corr:SGD}
From Theorem~\ref{theorem:SGD}, SGD has the iteration complexity of
\begin{align}
    i^o > {\color{black}\frac{\ln \frac{2||w^o - {w}_0||^2}{\epsilon}}{\min\big\{\frac{\epsilon\nu}{6\sigma^2}, \frac{\nu}{7\delta^2}  \big\}\cdot \nu} }.
\end{align}
\end{corollary}

\begin{corollary}[Iteration complexity of SVRG] \label{corr:SVRG}
From Theorem~\ref{theorem:SVRG}, SVRG has the iteration complexity of
{\color{black}\begin{align}
    i^o \geq \frac{2\ln \frac{||w^o - w_0||^2}{\epsilon}}{\min\big\{ \frac{\nu}{9\delta^2}, \frac{1}{m\nu} \big\}\big( \nu - \frac{16\delta^2}{m\nu} \big)} 
\end{align}
for any $m>16\delta^2/\nu^2$.}

\end{corollary}

\section{Convergence Analysis of $\alpha$-SVRG under Strong Convexity}  \label{sec:fixed}
{\color{black}We now study $\alpha$-SVRG \cite{yin2023coefficient}}.  The structure of the argument is analogous to that leading to Theorems~\ref{theorem:SGD} and~\ref{theorem:SVRG}. {\color{black} All the proofs are deferred to Sec.~\ref{sec:proofs}.}

\begin{lemma}[MSD bound of $\alpha$-SVRG gradient noise] \label{lemma:MSE_alpha_SVRG} The MSD between the risk gradient $\nabla J(w)$ and the $\alpha$-SVRG gradient $\widehat{\nabla J}^\text{$\alpha$-SVRG} = \alpha(\nabla J(\bar{w}) - \nabla Q(\bar{w};\x_i))+\nabla Q(w;\x_{i}) $ given the snapshot model $\bar{w}$ has the following upper bound for any fixed $w$ and \( \overline{w} \):
\begin{align}
    &\mathbb{E}||\nabla J(w) - \widehat{\nabla J}^\text{$\alpha$-SVRG}(w)||^2 \\&\leq  {\color{black}(2\alpha + 6)\delta^2}|| w^o - w ||^2 + 8{\color{black}\alpha} \delta^2 ||w^o - \bar{w}||^2  + 3{\color{black}(1-\alpha)}  \sigma^2\, \nonumber
\end{align}   
which reduces to Lemma~\ref{lemma:MSE_SGD} by setting $\alpha=0$; to Lemma~\ref{lemma:MSE_SVRG} by setting $\alpha=1$.
\end{lemma}

\begin{theorem}[Convergence of $\alpha$-SVRG] \label{theor:conv_alpha_SVRG}
    Under the choice of the learning rate $\mu < \min\{ \nu/({\color{black}2\alpha}+7)\delta^2 , 1/m\nu\}$; of the snapshot period $m > 16{\color{black}\alpha}\delta^2/\nu^2$, $\alpha$-SVRG admits 
    \begin{align} 
     \mathbb{E}||w^o - {\w}_{i}||^2 \leq \Big( (1-\mu\nu)^m &+ \frac{8\mu\delta^2{\color{black}\alpha}}{\nu} \Big)^k ||w^o - {w}_{0 }||^2 \nonumber\\&\quad+ \frac{3\mu\sigma^2{\color{black}(1-\alpha)}}{\nu-8\mu\delta^2 {\color{black}\alpha}},
    \end{align}
    for $i=k\cdot m$.
\end{theorem}
Theorem~\ref{theor:conv_alpha_SVRG} {\color{black} exactly} reduces to Theorem~\ref{theorem:SGD} by setting $\alpha=0, m=1, k=i$; to Theorem~\ref{theorem:SVRG} by setting $\alpha=1$.

\begin{corollary}[Iteration complexity of $\alpha$-SVRG] \label{corr:alpha_SVRG}
From Theorem~\ref{theor:conv_alpha_SVRG}, $\alpha$-SVRG with $\alpha \in [0,1]$ has the iteration complexity of {\color{black}
\begin{align} \label{eq:unified_IC}
    i^o \geq \frac{2\ln\big( 2||w^o-w_0||^2/\epsilon(1+\alpha) \big)}{ \min\big\{ \frac{\nu}{(2\alpha+7)\delta^2},\frac{1}{m\nu}, \frac{\nu}{8\delta^2\alpha + {6\sigma^2(1-\alpha)}/{\epsilon}} \big\} \Big(\nu - \frac{16\delta^2{\color{black}\alpha}}{m\nu}\Big)}
\end{align}
for any $m > 16\alpha\delta^2/\nu^2$.}
\end{corollary}
The iteration complexity expression (\ref{eq:unified_IC}) recovers that of SGD (Corollary~\ref{corr:SGD}) up to a constant factor of $2$ by setting $\alpha=0$ and $m=1$; exactly recovers that of SVRG (Corollary~\ref{corr:SVRG}) by setting $\alpha=1$.

\subsection{Take-Away}
The iteration complexity expression (\ref{eq:unified_IC}) enables efficient choice of the reliance factor \( \alpha \)  {\color{black} depending on the systems parameters at hand. We consider the following practically interesting cases: (\emph{i}) high stochasticity $\sigma^2 \gg 1$ and/or (\emph{ii}) low target error $\epsilon \ll 1$, for which the learning rate upper bound in (\ref{eq:unified_IC}) is dominated by the last term, i.e.,} 
\begin{align} \label{eq:alpha_trend}
    \mu < \frac{\nu}{8\delta^2\alpha + {6\sigma^2(1-\alpha)}/{\epsilon}}  = \frac{\nu}{\big( 8\delta^2 - \frac{6\sigma^2}{\epsilon}\big) \alpha + \frac{6\sigma^2}{\epsilon}}.
\end{align}
{\color{black} In above, we consider $m$ to be not excessively large so as to avoid excessively slow convergence rate by noting that the iteration complexity $i^o$ in (\ref{eq:unified_IC}) becomes directly proportional to $m$, i.e., $i^o \approx 2 m \ln ( 2||w^o-w_0||^2/\epsilon(1+\alpha) )$ with $m\gg 1$. }

{\color{black}It is clear from (\ref{eq:unified_IC}) that, in order to achieve low iteration complexity, one needs to choose as large learning rate as possible.} To this end, we focus on the trend of the term (\ref{eq:alpha_trend}) as a function of $\alpha$. Depending on the multiplicative term $(8\delta^2-6\sigma^2/\epsilon)$ being positive or negative, the upper bound decreases or increases with increased $\alpha$, respectively. Specifically, in order to increase (\ref{eq:alpha_trend}), we need to choose $\alpha$ as follows \begin{align}
\begin{cases}
    \text{decrease $\alpha$ if  $6\sigma^2 < 8\epsilon\delta^2$ (small stochasticity)}  \\ \text{increase $\alpha$ if $6\sigma^2 \geq 8\epsilon\delta^2$ (large stochasticity)},
    \end{cases}
\end{align}
which is aligned with the intuition that SVRG-type optimization becomes useful in the presence of large stochasticity, while SGD-type algorithm is preferable in low stochasticity.

\section{Proofs} \label{sec:proofs}
In this section, we provide all the essential proof details for our analysis in the previous section. 
\subsection{Proof of Lemma~\ref{lemma:MSE_alpha_SVRG}}
Proof is direct by considering the standard gradient noise bound for any fixed $w$ \cite{sayed2022inference}
\begin{align} \label{eq:gradient_noise_bound}
    \mathbb{E}[ || \nabla J(w) - \widehat{\nabla J}(w) ||^2] \leq 6\delta^2 ||w^o - w||^2 + 3\sigma^2
\end{align}
with the following inequalities given any fixed $w$ and $\bar{w}$:
\begin{align}
    &\mathbb{E}[||\nabla J(w) - \widehat{\nabla J}^\text{$\alpha$-SVRG}(w)||^2] =\mathbb{E}[|| \alpha(\nabla J(w) -\nabla J(\bar{w})) \nonumber\\& + (1-\alpha)(\nabla J(w) - \widehat{\nabla J}(w)) -\alpha(\widehat{\nabla J}(w) - \widehat{\nabla J}(\bar{w}) ) ||^2] \nonumber\\
    \overset{(a)}{\leq}& \frac{4\alpha^2\delta^2}{\color{black}{\beta}}||w-\bar{w}||^2 + \frac{(1-\alpha)^2}{{\color{black}1-\beta}} \Big( 6\delta^2 ||w^o - w||^2 + 3\sigma^2\Big)
    \nonumber\\\overset{(b)}{\leq}& (2\alpha+6)\delta^2||w^o-w||^2 + 8{\color{black}\alpha}\delta^2||w^o - \bar{w}||^2+ 3{\color{black}(1-\alpha)}\sigma^2,
\end{align}
where (a) uses  Jensen's inequality for any $\beta \in [0,1]$ along with Assumption~\ref{assump:Lip} and (\ref{eq:gradient_noise_bound}); (b) uses $||w_{i-1} - w^o + w^o - \bar{w}||^2 \leq 2||w - w^o||^2 + 2||w^o - \bar{w}||^2$ and by setting $\beta=\alpha$.

\subsection{Proof of Theorem~\ref{theor:conv_alpha_SVRG}}
After subtracting $w^o$ to both sides of the update rule (\ref{eq:svrg_update}), then by considering Lemma~\ref{lemma:MSE_alpha_SVRG} as well as Assumptions~\ref{assump:Lip} and \ref{assump:convexity} for a given $\w_{i-1}$ and $\bar{\w}_{i-1}$, we have 
\begin{align}
    &\mathbb{E}[||w^o - \w_i||^2|\w_{i-1}, \bar{\w}_{i-1}] \nonumber\\&\leq \big( 1  - 2\mu\nu + ({\color{black}2\alpha} + 7)\mu^2\delta^2 \big)||w^o - {\w}_{i-1}||^2\nonumber\\& + 8{\color{black}\alpha}\delta^2 \mu^2 ||w^o - \bar{\w}_{i-1}||^2+3{\color{black}(1-\alpha)}\sigma^2\mu^2. \label{eq:sangwoos_intermediate_result}
\end{align}
The statement follows after iterating and applying arguments analogous to those leading to Theorem~\ref{theorem:SVRG}~\cite{sayed2022inference} to account for the mismatch of the snapshot model\(  \).

Let us then denote as $A(\alpha) = 1 - 2\mu \nu + ({\color{black}2\alpha}+7)\mu^2\delta^2$; $B(\alpha) = 8{\color{black}\alpha}\delta^2\mu^2$; $C(\alpha) = 3{\color{black}(1-\alpha)}\sigma^2\mu^2$, with which (\ref{eq:sangwoos_intermediate_result}) can be rewritten by taking expectations over $\w_{i-1}$ and $\bar{\w}_{i-1}$
\begin{align}
    &\mathbb{E}||\tilde{\w}_{km + j}||^2 \nonumber\\&\leq A(\alpha)\mathbb{E}||\tilde{\w}_{km+j-1}||^2 + B(\alpha)\mathbb{E}||\tilde{\w}_{km}||^2 + C(\alpha),
\end{align}
where we denote as $\tilde{\w}_i = w^o - \w_i$ the error of the model $\w_i$.
    
By assuming $\mu \leq \nu/({\color{black}2\alpha}+7)\delta^2$ to ensure $A(\alpha) \leq 1-\mu\nu$, and to apply the standard recursion process (see, e.g., \cite{sayed2022inference}) over the inner iterations for a fixed outer epoch $k$, we get
\begin{align}
    &\mathbb{E}||\tilde{\w}_{(k+1)m}||^2 \nonumber\\&\leq  \underbrace{\bigg( (1-\mu\nu)^m + 8{\color{black}\alpha}\delta^2\mu \Big( \frac{1-(1-\mu\nu)^m}{\nu} \Big) \bigg)}_{:=\tilde{A}} \mathbb{E}||\tilde{\w}_{km}||^2\nonumber\\&\quad\quad\quad\quad\quad\quad\quad\quad\quad+ \underbrace{3{\color{black}(1-\alpha)}\mu\sigma^2\frac{1-(1-\mu\nu)^m}{\nu}}_{:=\tilde{B}}. 
\end{align}
Now by applying the recursion with respect to the outer epoch $k$, we get $\mathbb{E}||\tilde{\w}_{k\cdot m}||^2 \leq \tilde{A}^k ||\tilde{w}_{0}||^2 + {\tilde{B}}/{(1-\tilde{A})}.$

Imposing $\mu \leq 1/m\nu $ to ensure $(1-\mu\nu)^m \leq 1-\mu\nu m/2$ using Taylor expansion, we find the upper bound $\tilde{A}^\text{u}$ on $\tilde{A}$ as
\begin{align} \label{eq:A_u}
    \tilde{A}^\text{u} = 1-\frac{\mu\nu m}{2}+ 8{\color{black}\alpha}\delta^2\mu \Big( \frac{1}{\nu} \Big) \geq \tilde{A}.
\end{align}
It is now readily checked that
\begin{align}
    \mathbb{E}||\tilde{\w}_{k\cdot m}||^2 
    &\leq (\tilde{A}^\text{u})^k ||\tilde{w}_{0}||^2  +   \frac{\tilde{B}}{1-\tilde{A}^\text{u}}
\end{align}
which ensures stability condition $\tilde{A}\leq \tilde{A}^\text{u}< 1$ upon the choice of  $m>16{\color{black}\alpha}\delta^2/\nu^2$. 

Although we have used the upper bound $\tilde{A}^\text{u}$ to set the conditions for $\mu$ and $m$, we now go back to the tighter expression that uses $\tilde{A}$ to get the actual recursion expression
\begin{align}
    &\mathbb{E}||\tilde{\w}_{k\cdot m}||^2 \nonumber\\&\leq \bigg( (1-\mu\nu)^m + 8{\color{black}\alpha}\delta^2\mu\cdot \Big( \frac{1-(1-\mu\nu)^m}{\nu} \Big) \bigg)^k||\tilde{w}_0||^2 \nonumber\\&+ \frac{3{\color{black}(1-\alpha)}\mu\sigma^2\frac{1-(1-\mu\nu)^m}{\nu}}{1- (1-\mu\nu)^m - 8{\color{black}\alpha}\delta^2\mu\cdot \Big( \frac{1-(1-\mu\nu)^m}{\nu} \Big) }\nonumber\\
    &  \leq \Big( (1-\mu\nu)^m + \frac{8{\color{black}\alpha}\delta^2\mu}{\nu} \Big)^k||\tilde{w}_0||^2 + \frac{3{\color{black}(1-\alpha)}\mu\sigma^2}{\nu-8{\color{black}\alpha}\delta^2\mu}
\end{align}
where the last inequality uses $1-(1-\mu\nu)^m \leq 1$ from $\mu\nu \in (0,1)$ {\color{black} (first drift term}); also to cancel out $1-(1-\mu\nu)^m $ from the numerator and the denominator {\color{black}(second stochasticity term). }

\subsection{Proof of Corollary~\ref{corr:alpha_SVRG}}
From Theorem~\ref{theor:conv_alpha_SVRG}, we consider the following sufficient condition to achieve the target performance level $\epsilon$, i.e., 
\begin{subequations}
\begin{align} \label{eq:first_iter}
     \Big( (1-\mu\nu)^m + \frac{8\mu\delta^2{\color{black}\alpha}}{\nu} \Big)^k ||\tilde{w}_{0 }||^2  &\leq \frac{\epsilon}{2};\\
     \frac{3\mu\sigma^2{\color{black}(1-\alpha)}}{\nu-8\mu\delta^2 {\color{black}\alpha}} &\leq \frac{\epsilon}{2}. \label{eq:second_iter}
\end{align}
\end{subequations}
First, from the second condition (\ref{eq:second_iter}), we set the learning rate{\color{black}  
\begin{align} \label{eq:mu_iter_alpha}
    \mu \leq \frac{\nu}{8\delta^2\alpha + {6\sigma^2(1-\alpha)}/{\epsilon}}.
\end{align}} {\color{black} We then turn to the} first condition (\ref{eq:first_iter}) to get the epoch complexity $k^o$ by recalling the useful inequality $1-x \leq e^{-x}$ for all $x\in\mathbb{R}$, i.e., consider the sufficient condition for (\ref{eq:first_iter})
\begin{align} \label{eq:epoch_complexity}
     e^{k\cdot \big(-\frac{\mu\nu m}{2} +\frac{8\mu\delta^2\alpha}{\nu}\big)} \leq \frac{\epsilon}{2}
\end{align}
where we have used $(1-\mu\nu)^m \leq 1-\mu\nu m/2$ again as per (\ref{eq:A_u}).
Solving (\ref{eq:epoch_complexity}), we get \begin{align} \label{eq:epoch_complexity_2}
    k^o \geq \frac{\ln\big( 2||\tilde{w}_0||^2/\epsilon \big)}{\mu\Big(\frac{1}{2}m\nu - \frac{8\delta^2{\color{black}\alpha}}{\nu}\Big)},
\end{align}
which monotonically decreases with increasing $\mu$. Thus, it is beneficial to choose the maximal learning rate $\mu$. Combining the new constraint (\ref{eq:mu_iter_alpha}) as well as the existing constraint $\mu < \min\{ \nu/({\color{black}2\alpha}+7)\delta^2 , 1/m\nu\}$, we have the complete constraint on $\mu$
\begin{align}
    \mu \leq \min\bigg\{ \frac{\nu}{(2\alpha+7)\delta^2}, \frac{1}{m\nu}, {\color{black}\frac{\nu}{8\delta^2\alpha + {6\sigma^2(1-\alpha)}/{\epsilon}}} \bigg\}
\end{align} {\color{black}
from which we can write the epoch complexity as\begin{align} \label{eq:epoch_complexity_3}
    k^o \geq \frac{2\ln\big( 2||\tilde{w}_0||^2/\epsilon(1+\alpha) \big)}{ \min\big\{ \frac{\nu}{(2\alpha+7)\delta^2},\frac{1}{m\nu}, \frac{\nu}{8\delta^2\alpha + {6\sigma^2(1-\alpha)}/{\epsilon}} \big\} \Big(m\nu - \frac{16\delta^2{\color{black}\alpha}}{\nu}\Big)}.
\end{align}
Corollary~\ref{corr:alpha_SVRG} is now direct from the relationship $i^o = m\cdot  k^o$.
}



\begin{figure} [t]
    \centering
    \includegraphics[width=0.8\linewidth]{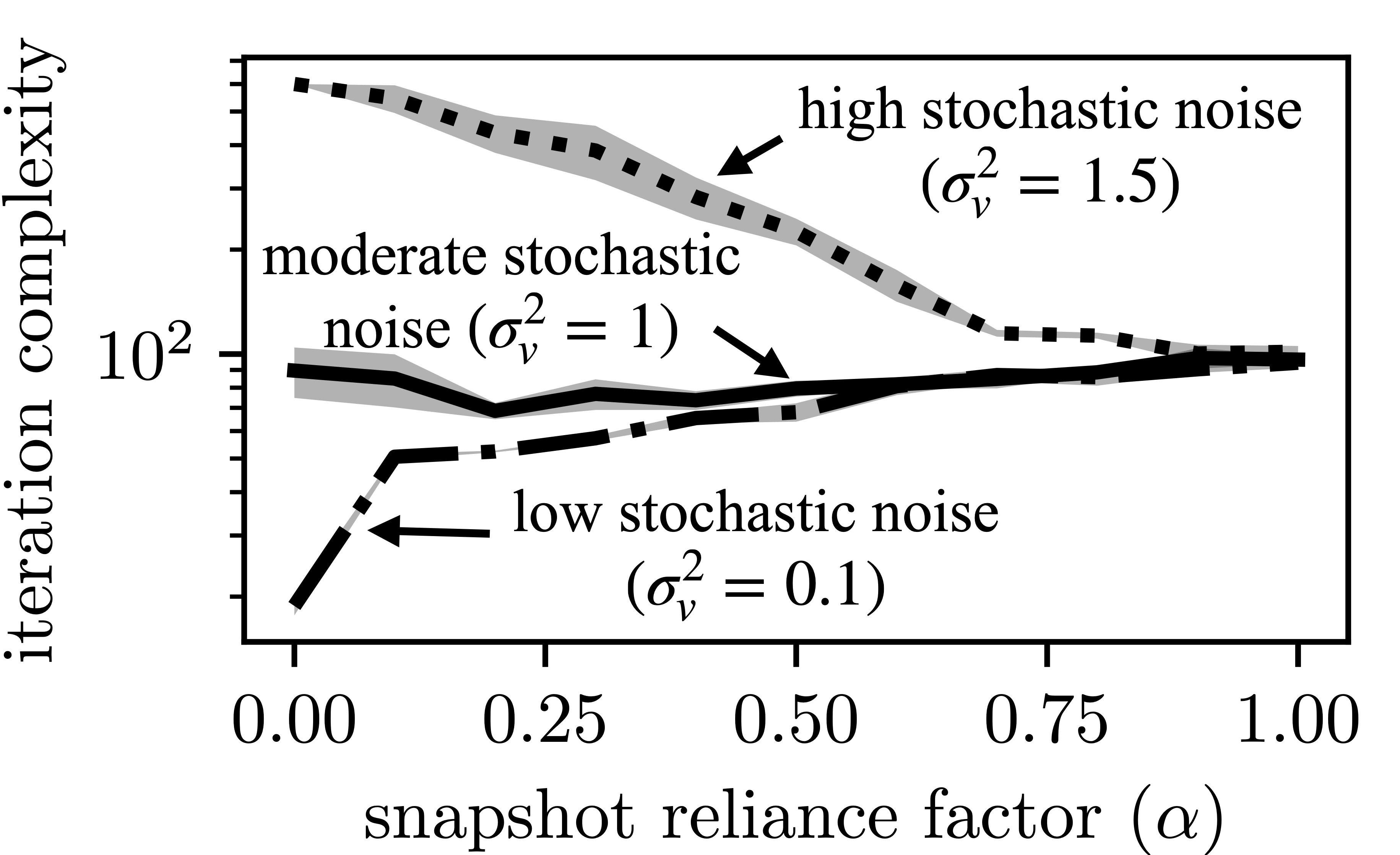}
    \caption{Iteration complexity (\ref{eq:def_iteration_complexity}) as a function of $\alpha$ for different levels of observation noise. Shaded area corresponds to $95\%$ confidence interval obtained from $10$ independent experiments.}
    \label{fig: Exp_noise}
\end{figure}

\section{Numerical Results}
We consider the linear regression problem where the label variable $\boldsymbol{\gamma} \in\mathbb{R}$ is generated by
\begin{equation} \label{eq:linear_regression}
    \boldsymbol{\gamma} = \boldsymbol{h}^\top w^o + \boldsymbol{v}  
\end{equation}
given the feature $\boldsymbol{h}\in\mathbb{R}^2$ and the ground-truth linear vector $w^o \in \mathbb{R}^2$, with Gaussian observation noise $v \sim \mathcal{N}(0,\sigma_v^2)$.
Given $x_n = (h_n,\gamma_n)$ obtained from (\ref{eq:linear_regression}) for $n=1,...,N$, the risk is defined as $J(w) = \frac{1}{N} \sum_{n=1}^N |h_n^\top w - \gamma_n|^2$.

We set the total number of samples $N=50$; inner iteration size $m=50$; and target MSE $\epsilon = 5\cdot 10^{-3}$. In order to evaluate the iteration complexity (\ref{eq:def_iteration_complexity}), we first estimate the expected MSD for different learning rates ranging from $0.05$ to $0.6$ to choose the minimum number of iterations required to reach the target $\epsilon$. We estimate the expected MSE by averaging over $10$ independent runs. 

By noting that  $\sigma^2 = (2/N)\sum_{n=1}^N ||h_n v_n||^2$ is proportional to the observation noise variance $\sigma_v^2$, we consider the following three cases with (\emph{i}) low observation noise $\sigma_v^2 = 0.1$; (\emph{ii}) moderate observation noise $\sigma_v^2=1$; and (\emph{iii}) large observation noise $\sigma_v^2 = 1.5$. In Fig.~\ref{fig: Exp_noise}, we plot the iteration complexity $i^o$ as a function of $\alpha$.

As predicted by Theorem~\ref{theor:conv_alpha_SVRG}, in the large stochastic gradient variance regime, SVRG is preferable to SGD; while smaller stochastic gradient variance allows SGD to perform better than SVRG. In the moderate observation noise regime, $\alpha$-SVRG performs better than both SGD and SVRG with $\alpha=0.2$, supporting Corollary~\ref{corr:alpha_SVRG}.

\section{Conclusion}
In this work, we provided analytical arguments for the success of a recently proposed variant of SVRG, named $\alpha$-SVRG \cite{yin2023coefficient}, in strongly convex environments. Theoretical results on both fixed and time-varying $\alpha$ suggests the choice of $\alpha$ proportional to the variance $\sigma^2$ of the gradient evaluated at the optimal model $w^o$, which is also supported by numerical results in linear regression. {\color{black} Future work may extend the analysis of $\alpha$-SVRG to introduce the hyperparameter $\alpha$ to the other variance-reduction techniques such as SAGA \cite{Defazio14}, SARAH \cite{nguyen2017sarah}, and SPIDER \cite{fang2018spider}. }

\bibliographystyle{IEEEtran}
\bibliography{ref.bib}

\begin{thebibliography}{10}
\providecommand{\url}[1]{#1}
\csname url@samestyle\endcsname
\providecommand{\newblock}{\relax}
\providecommand{\bibinfo}[2]{#2}
\providecommand{\BIBentrySTDinterwordspacing}{\spaceskip=0pt\relax}
\providecommand{\BIBentryALTinterwordstretchfactor}{4}
\providecommand{\BIBentryALTinterwordspacing}{\spaceskip=\fontdimen2\font plus
\BIBentryALTinterwordstretchfactor\fontdimen3\font minus \fontdimen4\font\relax}
\providecommand{\BIBforeignlanguage}[2]{{%
\expandafter\ifx\csname l@#1\endcsname\relax
\typeout{** WARNING: IEEEtran.bst: No hyphenation pattern has been}%
\typeout{** loaded for the language `#1'. Using the pattern for}%
\typeout{** the default language instead.}%
\else
\language=\csname l@#1\endcsname
\fi
#2}}
\providecommand{\BIBdecl}{\relax}
\BIBdecl

\bibitem{yin2023coefficient}
Y.~Yin, Z.~Xu, Z.~Li, T.~Darrell, and Z.~Liu, ``A coefficient makes svrg effective,'' \emph{arXiv preprint arXiv:2311.05589}, 2023.

\bibitem{sayed2022inference}
A.~H. Sayed, \emph{Inference and Learning from Data: Learning}.\hskip 1em plus 0.5em minus 0.4em\relax Cambridge University Press, 2022, vol.~3.

\bibitem{simeone2022machine}
O.~Simeone, \emph{Machine learning for engineers}.\hskip 1em plus 0.5em minus 0.4em\relax Cambridge university press, 2022.

\bibitem{Yuan16}
K.~Yuan, B.~Ying, S.~Vlaski, and A.~H. Sayed, ``Stochastic gradient descent with finite samples sizes,'' in \emph{Proc. IEEE MLSP}, 2016, pp. 1--6.

\bibitem{ross2022simulation}
S.~M. Ross, \emph{Simulation}.\hskip 1em plus 0.5em minus 0.4em\relax academic press, 2022.

\bibitem{johnson2013accelerating}
R.~Johnson and T.~Zhang, ``Accelerating stochastic gradient descent using predictive variance reduction,'' \emph{Advances in neural information processing systems}, vol.~26, 2013.

\bibitem{Defazio14}
A.~Defazio, F.~Bach, and S.~Lacoste-Julien, ``{SAGA: A} fast incremental gradient method with support for non-strongly convex composite objectives,'' in \emph{Advances in Neural Information Processing Systems}, vol.~27, 2014.

\bibitem{zhang2013linear}
L.~Zhang, M.~Mahdavi, and R.~Jin, ``Linear convergence with condition number independent access of full gradients,'' \emph{Advances in Neural Information Processing Systems}, vol.~26, 2013.

\bibitem{mahdavi2013mixed}
M.~Mahdavi, L.~Zhang, and R.~Jin, ``Mixed optimization for smooth functions,'' \emph{Advances in neural information processing systems}, vol.~26, 2013.

\bibitem{wang2013variance}
C.~Wang, X.~Chen, A.~J. Smola, and E.~P. Xing, ``Variance reduction for stochastic gradient optimization,'' \emph{Advances in neural information processing systems}, vol.~26, 2013.

\bibitem{nguyen2017sarah}
L.~M. Nguyen, J.~Liu, K.~Scheinberg, and M.~Tak{\'a}{\v{c}}, ``Sarah: A novel method for machine learning problems using stochastic recursive gradient,'' in \emph{International conference on machine learning}.\hskip 1em plus 0.5em minus 0.4em\relax PMLR, 2017, pp. 2613--2621.

\bibitem{fang2018spider}
C.~Fang, C.~J. Li, Z.~Lin, and T.~Zhang, ``Spider: Near-optimal non-convex optimization via stochastic path-integrated differential estimator,'' \emph{Advances in neural information processing systems}, vol.~31, 2018.

\bibitem{wang2019spiderboost}
Z.~Wang, K.~Ji, Y.~Zhou, Y.~Liang, and V.~Tarokh, ``Spiderboost and momentum: Faster variance reduction algorithms,'' \emph{Advances in Neural Information Processing Systems}, vol.~32, 2019.

\bibitem{reddi2016stochastic}
S.~J. Reddi, A.~Hefny, S.~Sra, B.~Poczos, and A.~Smola, ``Stochastic variance reduction for nonconvex optimization,'' in \emph{International conference on machine learning}.\hskip 1em plus 0.5em minus 0.4em\relax PMLR, 2016, pp. 314--323.

\bibitem{frostig2015competing}
R.~Frostig, R.~Ge, S.~M. Kakade, and A.~Sidford, ``Competing with the empirical risk minimizer in a single pass,'' in \emph{Conference on learning theory}.\hskip 1em plus 0.5em minus 0.4em\relax PMLR, 2015, pp. 728--763.

\bibitem{lei2017less}
L.~Lei and M.~Jordan, ``Less than a single pass: Stochastically controlled stochastic gradient,'' in \emph{Artificial Intelligence and Statistics}.\hskip 1em plus 0.5em minus 0.4em\relax PMLR, 2017, pp. 148--156.

\bibitem{cutkosky2019momentum}
A.~Cutkosky and F.~Orabona, ``Momentum-based variance reduction in non-convex sgd,'' \emph{Advances in neural information processing systems}, vol.~32, 2019.

\bibitem{Defazio19}
A.~Defazio and L.~Bottou, ``On the ineffectiveness of variance reduced optimization for deep learning,'' in \emph{Advances in Neural Information Processing Systems}, vol.~32, 2019.

\end{thebibliography}

\end{document}